\documentclass{article}

\usepackage{PRIMEarxiv}
\usepackage[pdftex,
            pdfauthor={J.Rafid Siddiqui},
            pdftitle={FExGAN-Meta: Facial Expression Generation with Meta Humans},
            pdfsubject={Computer Vision},
            pdfkeywords={"Facial-Expressions, Deep Learning, Generative Adversarial Network, Computer Vision"},
            pdfproducer={Latex with hyperref, or other system},
            pdfcreator={pdflatex, or other tool}]{hyperref}
\usepackage[utf8]{inputenc} 
\usepackage[T1]{fontenc}    
\usepackage{hyperref}       
\usepackage{url}            
\usepackage{booktabs}       
\usepackage{amsfonts}       
\usepackage{nicefrac}       
\usepackage{microtype}      
\usepackage{lipsum}
\usepackage{fancyhdr}       
\usepackage{graphicx}       
\graphicspath{{./images/} }    
\usepackage{hyperref}
\usepackage{bm}			
\usepackage{floatrow}	
\newfloatcommand{capbtabbox}{table}[][\FBwidth]

\title{FExGAN-Meta: Facial Expression Generation with Meta Humans
}

\author{
  J. Rafid Siddiqui \\
  \\
  AZAD Research Lab \\
  Sweden\\
  \texttt{jrs@azaditech.com} 
}

\begin{document}
\maketitle

\begin{abstract}
The subtleness of human facial expressions and a large degree of variation in the level of intensity to which a human expresses them is what makes it challenging to robustly classify and generate images of facial expressions. Lack of good quality data can hinder the performance of a deep learning model. In this article, we have proposed a Facial Expression Generation method for Meta-Humans (FExGAN-Meta) that works robustly with the images of Meta-Humans. We have prepared a large dataset of facial expressions exhibited by ten Meta-Humans when placed in a studio environment and then we have evaluated FExGAN-Meta on the collected images. The results show that FExGAN-Meta robustly generates and classifies the images of Meta-Humans for the simple as well as the complex facial expressions.

\end{abstract}

\keywords{Facial-Expressions \and Deep Learning \and Generative Adversarial Network \and Computer Vision}

\section{Introduction}

The power of effective communication has lead humans reach the top of the food chain. Humans do not posses the physical might of other animals however, this didn't limit humans from ruling the planet. In addition to gathering the knowledge about the environment, humans have developed tools and systems in order to transfer the learning from one person to another. This has lead to the exponential growth of knowledge. Humans have devised a diverse set of tools (e.g. speech, facial expressions, body language, written/typed text, videos, illustrations, paintings, 3D visualizations etc.). Some of these tools are primitive and are built into the subconscious of a human brain due to them being the input source of a brain during the brain forming years. Facial expressions, along with body language is one of these tools that humans use frequently and subconsciously as a primary mode of communication. \\

The subtleness of human facial expressions and a large degree of variation in the level of intensity to which a human expresses them is what makes it challenging to robustly classify and generate images of facial expressions. Further, there also exists variation among different identities expressing the same emotion. These subtleties and variations are often lost or not captured properly while preparing a dataset that is intended to be used for facial expressions due to the limitations regarding the collection of data from real subjects. In the past few years, computer graphics research has advanced to a point where the photo-realism of artificial characters has become possible. These artificial characters are called Meta-Humans \cite{metahuman}. These Meta-Humans are rendered to the level of individual hair strand, in contrast to the earlier attempts where meshes were a common practice in order to reduce the overall computation. The resulting characters are indistinguishable to the naked eye and thus provide a rich source of data for training learning models specially the deep-nets. The model trained on Meta-humans can provide a basis for zero-shot or a few shot transfer to the real images. As the progress in the development of these characters continues, it should soon be possible to directly apply the model trained on meta-human to real-human images.    \\

In addition to the data, a learning framework requires a robust model. In the past several decades, face recognition has been the important subject of research.  In the recent years, it has become a task that can robustly be performed by the state-of-the-art methods \cite{parkhi_deep_2015}\cite{cao_vggface2_2018}. However, facial expression recognition is still an active research topic with some success over the past years. Recently, facial expression generation has emerged as a new field of research which deals with not only the classification problem but also allows exploration of the facial expression space. This is much more challenging task however, it provides with better understanding of the underlying semantics of the facial expression. The resulting system not only provides a labeling system but tries to map the space therefore, understands the data at deeper level. This has many applications in real world, such as video-editing, animation and game development being couple of them. \\

The concept of a Generative model was proposed in the form of a GAN architecture \cite{goodfellow_generative_2014}. It is a network architecture that generates images from a random Gaussian noise. A large number of research work has appeared in recent years regarding Generative Adversarial Network (GAN) and has contributed in the advancement of computer vision research. 
A classical GAN network consists of two networks: a generator and a discriminator network. The generator network is a network that takes a random sample drawn from a Gaussian distribution, called the latent vector, and reconstructs an image. The discriminant network takes  the real image  and the false output from the generator network and classifies it as true or false. The GAN architecture with Convolutional layers as well as the dense layers was proposed later and was called Deep Convolutional GAN (DCGAN)\cite{radford_unsupervised_2016}. The classical GAN architecture does not have any control over the type of output from the generator, however, a conditional GAN (cGAN) has input in the form of a class label and can produce images of a particular class \cite{mirza_conditional_2014}. A cGAN architecture can generate multiple different kind of images with different class labels however, it the discriminator in the architecture still produces a binary output of whether an image is real or a fake one. In order to extend the concept of cGAN for handling multiple classes in the discriminator network, a Auxiliary Classifier (AC-GAN) was proposed. AC-GAN has a discriminator network which gives a binary decision as well as the class label as an output. \cite{odena_conditional_2017}. In an earlier article \cite{siddiqui_explore_2022}, we introduced a GAN architecture that takes inspiration from AC-GAN as well as autoencoder network, and proposed FExGAN which could generate multiple classes of expressions robustly of cartoon characters. In this article, we propose FExGAN-Meta which can robustly generate images of Meta-Humans robustly.

The rest of the article is organized in the following manner. In section \ref{sec:related-work} we list closest work in the field about facial expressions. In section \ref{sec:method} we list the FExGAN-Meta architecture. In section \ref{sec:experimentation} we elaborate the process for the preparation of meta-human dataset and the training procedure. In section \ref{sec:results} we present the results of the method applied on meta-human dataset and in section \ref{sec:conclusions} we conclude our work.

\section{Related Work}
\label{sec:related-work}

Image generation and facial image generation in particular is a topic that has its origin in the recent years. Therefore, the most relevant work exists is the one that addresses facial expression recognition. There is significant body of work in Facial expression recognition domain, however, the most relevant work would be that which solves the problem using a deep learning approach. In \cite{kuo_compact_2018} a facial expression recognition method was proposed and evaluated on multiple image datasets. In \cite{khaireddin_facial_2021} a CNN based architecture was introduced which achieves state-of-the-art accuracy over FER2013 dataset with parameter optimization. In \cite{sajjanhar_deep_2018} another CNN based architecture was proposed and evaluated on images of faces and identities were compared using pre-trained face model \cite{cao_vggface2_2018}. In  \cite{ning_emotiongan_2020} a GAN based network was constructed and which projects the image into latent space and then a Generalized Linear Model (GLM) was used to fit the direction of various facial expressions. A GAN architecture was proposed in \cite{deng_cgan_2019} which could classify and generate images of a facial expression. A disentangled GAN network based architecture was proposed in \cite{ali_facial_2019} which could generate image of desired expressions. 

\section{Method}
\label{sec:method}

In this section we detail the proposed Facial Expression GAN for Meta-Humans(FExGAN-Meta) architecture. Then we explain the construction of the cost function, data-set construction and the training setup used for experiments. 

\subsection{FExGAN-Meta Architecture}

\begin{figure}[!htb]
  \centering
  \includegraphics[scale=.5]{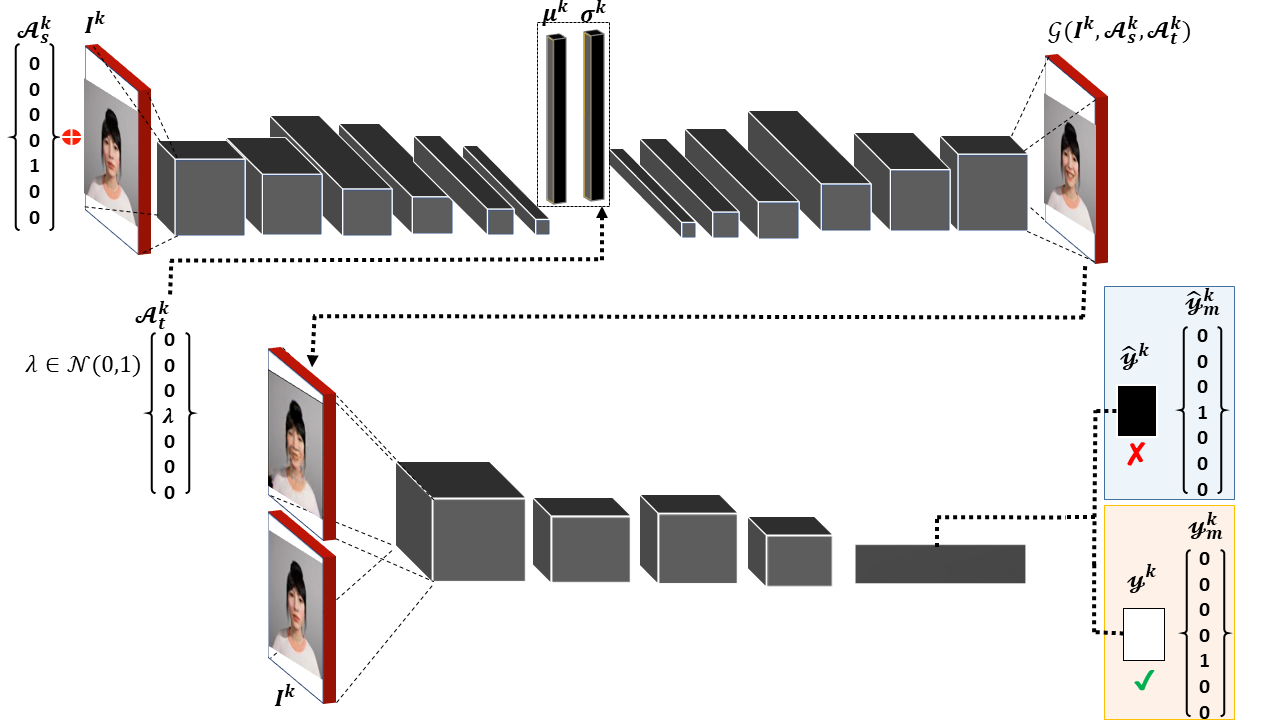}
  \caption{FExGAN Architecture with Meta-Humans}
  \label{fig:fexgan}
\end{figure}

A classical GAN architecture takes an input noise vector and results an image by applying a series of up-sampling layers. The images of a character identity in the training-set has a particular distribution which can be modeled as a normal distribution. Therefore, for an image  $\mathcal{I}^k$ of each character identity is projected into the latent space $\mathcal{Z}(\mathcal{I}^k) , \mathcal{Z} \subset \mathcal{N}(\mu^k,\sigma^k)$ which is sampled from a normal distribution. We shall make the generator learn this distribution by projecting an image into the latent space vector which shall be sampled from a learned normal distribution. The general architecture of the network can be seen in figure \ref{fig:fexgan}. \\

\par
 
In order to encode the input image of a facial expression, we shall input the condition $\mathcal{A}_s^k=[0,0,0,0,1,0,0]$ (i.e. an affect exhibited by the character) in addition to an input image. In addition to the modeling of the data distribution of a character, we would like to have control over the variations within a particular expression exhibited by a character. Since a sparse vector is not sufficient to gain control over the variations, therefore, we model each affect/expression $\mathcal{A}_t^k(\lambda) , \lambda \subset \mathcal{N}(0,1)$ randomly sampled from normal distribution. This sampled vector is then injected into the decoder along with the sampled latent vector obtained as an output from the encoder. This provides a fine-grained control over the individual expression exhibited by each character. \\

\subsection{Generator Architecture}

\begin{figure}[!htb]
  \centering
  \includegraphics[scale=.35]{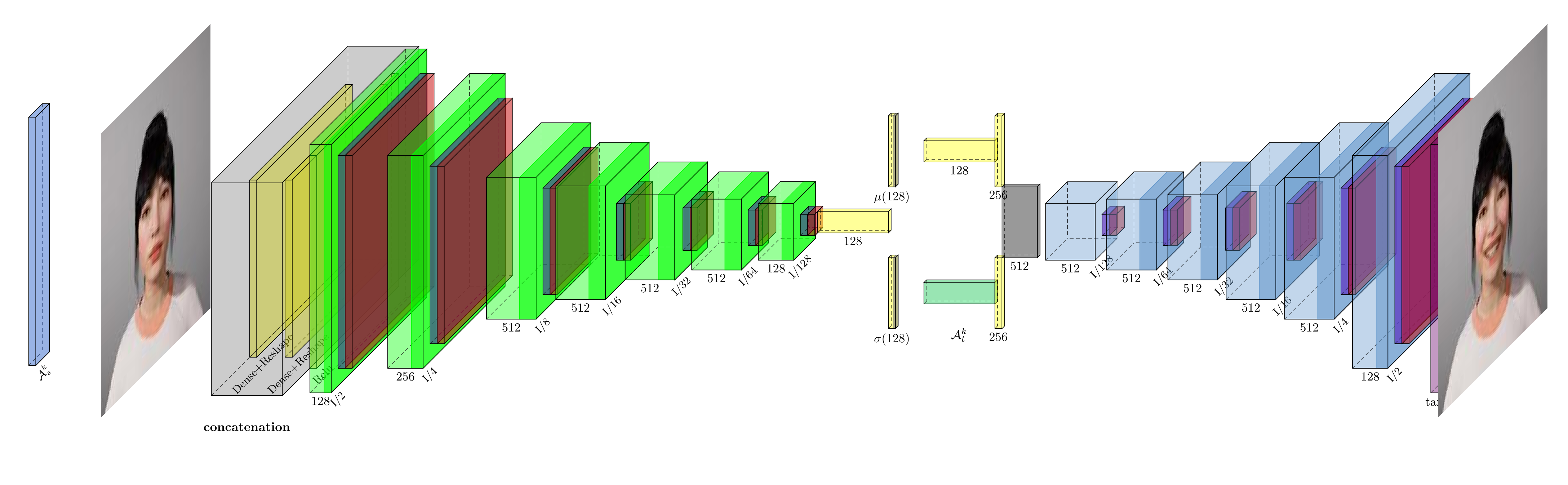}
  \caption{Layers in the Generator Network}
  \label{fig:generator}
\end{figure}

The generator network in the FExGAN-Meta architecture has an encoder and a decoder network. The encoder network projects an input image to a latent space $\mathcal{Z}(\mathcal{I}^k) , \mathcal{Z} \subset \mathcal{N}(\mu^k,\sigma^k)$ by learning the distribution of the data. In the encoder network there are several blocks of down-sampling layers. Each down-sampling block has three distinct layers: Convolution layer, a batch-normalization layer and an activation function layer. The 2D convolutions with strides has been used in order to perform convolution and down-sample in one step. The output of the convolution is batch-normalized and a linear activation is applied. There are a number of down-sampling blocks which are repeated until the image becomes uni-dimensional. Initial layers have increasing number of features while following layers consists of a fixed number of features. The output of the last layer in the decoder is flattened and attached to a fully connected set of layers: , $\mu$ and $\sigma$ which have n dimensions each. These layers provide the parameters for the latent space distribution. A latent vector is then sampled from the outputs of $\mu$ and $\sigma$ layers.  The layers of the encoder network can be visualized in the figure \ref{fig:generator}. \\

The latent vector formed as a result of the encoder output becomes an input of the decoder network along with the affect vector $\mathcal{A}_t^k(\lambda) , \lambda \subset \mathcal{N}(0,1)$. Another set of dense layers are attached and which are further concatenated and reshaped. There are six up-sampling blocks each consisting of 2D transpose convolutions, batch-normalization and activation layers. The conv2D transpose with strides achieves the opposite of conv2D with strides. The output if the up-sampling blocks generate half the size of the image and the last layer generates the full image with tanh as the activation function. \\

\subsection{Discriminator Architecture}

\begin{figure}
  \centering
  \includegraphics[scale=.5]{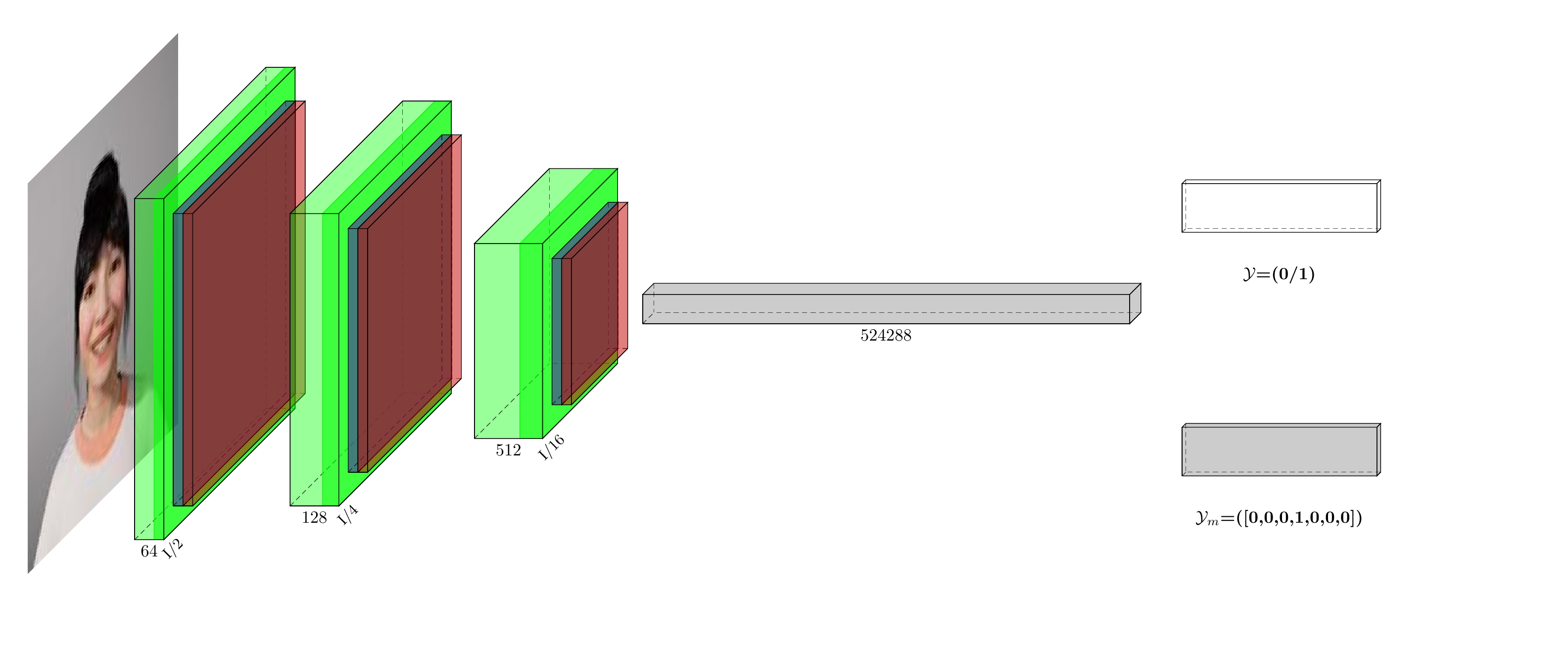}
  \caption{Layers in the Discriminator Network}
  \label{fig:discriminator}
\end{figure}

The task of the discriminator is to classify an input image into a set of affect classes in addition to determining the validity of the image. There are three blocks in the discriminator network and each block consists of convolutions, batch-normalization and activation layers. The number of features in each block are increased gradually and then the layers are flattened. A fully connected dense network is placed at the output of the network which consists of a sigmoid and a softmax layer responsible for the binary real/fake decision and the class label prediction respectively. The layers in the discriminator network can be seen in figure\ref{fig:discriminator}.

\subsection{Loss Function}

The central loss function of a GAN architecture is the adversarial loss. However, since generator in the proposed GAN architecture consists of a encoder-decoder networks therefore, we shall also include a reconstruction loss. The 'Adv' loss in the generator network for $\mathcal{I}^k$ is given in equation \ref{eq:eq1}.

\begin{equation}
\label{eq:eq1}
\mathcal{L}^\mathcal{G}_{Adv}(\mathcal{I}_g^k,\mathcal{A}_t^k,\hat{y}^k,\hat{y}_m^k) = \phi_b(\bm{1},\hat{y}^k) + \phi_m(\mathcal{A}_t^k,\hat{y}_m^k)
\end{equation}

where $\hat{y}^k,\hat{y}_m^k = \mathcal{D}(\mathcal{I}_g^k)$, $\phi_b$ and $\phi_m$ are the outputs of the discriminator $\mathcal{D}$ on fake image $\mathcal{I}^k_g=\mathcal{G}(\mathcal{I}^k_s,\mathcal{A}_s^k,\mathcal{A}_t^k )$ generated by generator $\mathcal{G}$, binary cross entropy and multi-class entropy functions respectively. \\

The generator network learns to produce not the the character displayed in the input image but also transform the facial expression supplied as a target affect $\mathcal{A}_t^k$. Therefore, it is needed that the reconstruction minimizes the difference between the generated image $\mathcal{I}^k_g$ and the target image $\mathcal{I}^k_t$. The target image is randomly selected from a list of images of the desired facial expression while keeping the character the same. The reconstruction loss is the L1-loss between the target and the real image and is given in equation \ref{eq:eq2}. \\

\begin{equation}
\label{eq:eq2}
\mathcal{L}_{reconst}(\mathcal{I}^k_s,\mathcal{I}^k_t,\mathcal{A}_s^k,\mathcal{A}_t^k) = \sum|\mathcal{G}(\mathcal{I}^k_s,\mathcal{A}_s^k,\mathcal{A}_t^k )-\mathcal{I}^k_t|
\end{equation}

As discussed in the earlier section, the distribution of each character identity is considered to be sampled from normal distribution, therefore, a KL-divergence between the learned and the unit-normal distribution is computed and is given in the equation \ref{eq:eq3}.

\begin{equation}
\label{eq:eq3}
\mathcal{L}_{KL}(\mu^k_i,\sigma^k_i) = \frac{-1}{2n} \sum_{i}^n[\bm{1}+\ln(\sigma^k_i)^2 -(\mu^k_i)^2 - e^{\ln(\sigma^k_i)^2} ] 
\end{equation}

The total loss of the generator is the combined loss consisting of the Adversarial ,reconstruction and the KL-divergence loss and is given in the equation \ref{eq:eq4}.

\begin{equation}
\label{eq:eq4}
\mathcal{L}_{gen} = \alpha \mathcal{L}^\mathcal{G}_{Adv} + \beta \mathcal{L}_{KL} + \gamma \mathcal{L}_{reconst} 
\end{equation}

The discriminator loss is computed for the real and the fake images separately and is given in the equation \ref{eq:eq5} and \ref{eq:eq6} respectively.

\begin{equation}
\label{eq:eq5}
\mathcal{L}_{real}(\bm{\mathcal{A}_s^k,y^k,y_m^k}) = \phi_b(\bm{1,y^k}) + \phi_m(\bm{\mathcal{A}_s^k,y_m^k})
\end{equation}

\begin{equation}
\label{eq:eq6}
\mathcal{L}_{fake}(\bm{\mathcal{A}_t^k,\hat{y}^k,\hat{y}_m^k}) = \phi_b(\bm{0,\hat{y}^k}) + \phi_m(\bm{\mathcal{A}_t^k,\hat{y}_m^k})
\end{equation}

where $y^k,y_m^k = \mathcal{D}(\mathcal{I}^k)$ is the output of the discriminator on real images. The discriminator loss then becomes the combined loss on the real and the fake images and is given in equation \ref{eq:eq7}.

\begin{equation}
\label{eq:eq7}
\mathcal{L}^\mathcal{D}_{Adv} = \mathcal{L}_{real} + \mathcal{L}_{fake}
\end{equation}

\section{Experimentation}
\label{sec:experimentation}
In this section, we describe the process of data preparation, model training and experimentation. \\

\subsection{Data Preparation}

\begin{figure}[!htb]
  \centering
  \includegraphics[scale=.4]{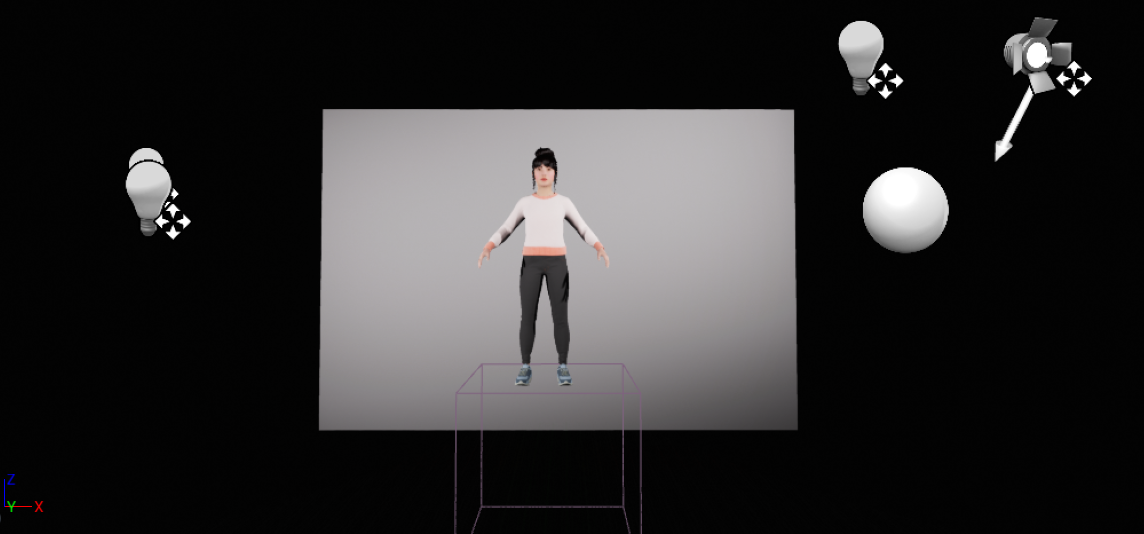}
  \caption{Studio Setup for Collection of Data with Meta-Humans}
  \label{fig:studio}
\end{figure}

We have prepared a Meta-Human Facial Expression Dataset (MH-FED) in this work. The Meta-Human characters are accessible through Meta-Human creator which is an online interface for generation and manipulation of Meta-Humans \cite{metahuman}. Firstly, we exported Meta-Humans from the creator to the Unreal Engine 4 (UE4). In order to have a standardized lightning setup for every Meta-Human imported into UE4, we have created a studio environment with careful placing of lightning to avoid shadows and over-exposure. A vertical plane is placed in order to provide neutral background and even spreading of light. Then each Meta-Human has been placed at a fixed distance from the plane and camera is zoomed to the face. The studio setup is shown in the figure \ref{fig:studio}. \\
In order for the character to perform a certain expression, we need to animate the character. Therefore, we have created a set of animations, one for each affect, and attached each animation to the character's face. Then when an animation is played the character exhibited that particular expression. All animations are then played sequentially for every character and a video output is recorded. The videos are then post-processed by splitting and cropping. There are ten characters that participated in the photo-shoot for a set of seven different affects, therefore, the dataset consists of 70 videos with each at \textasciitilde 35s@60fps. The frames in the videos are also exported in the form of images arranged in the directories for the convenience of data processing. The dataset provides \textasciitilde 162K images of size 512x512 with \textasciitilde  2300 images for each affect and character identity pair. This provides a rich source of data for learning the model for facial expressions.\\ 
The images in the dataset are processed with a set of operations in order to create a standard data for learning. More specifically, we resize and normalize each image in the dataset. The processed images are split into training and test sets. The training set consists of 113.4K images and the validation set consists of 48.6K respectively. \\
	
\subsection{Model Training}

The generator and discriminator network models are trained separately. However, the generator and the discriminator model work antagonistically so, they both learn the distribution of the data at convergence. Each training iteration consists of computation of the losses and then applying the loss on the network weights. In all experiments the learning rate of \emph{2e-4} has been used with \emph{adam} as an optimization method. In each step, a batch of target images are selected randomly for each character identity in the input image batch and the reconstruction loss is computed. Each training step consisted of a batch of images with a batch of 32 has been used in all experiments. The training is performed in batches with a batch size of 32 and process is iterated for 100K steps. The memory footprint of the model is \textasciitilde 7GB and the training process took \textasciitilde 2 weeks on a GTX1070 GPU with 8GB memory. \\

\begin{figure}
\begin{floatrow}
\ffigbox{%
\includegraphics[scale=0.5]{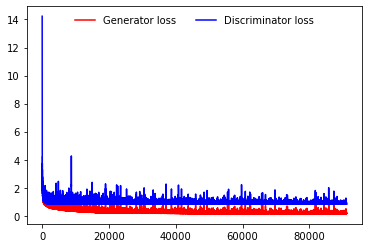}
}{
\caption{Convergence of loss function}%
}
\label{fig:loss}
\capbtabbox{%

\label{tab:accuracy}
  \centering
  \begin{tabular}{ccccc}
    \toprule

    & \multicolumn{2}{c}{Real}  & \multicolumn{2}{c}{Fake}                 \\
    \cmidrule(r){2-3}    
	\cmidrule(r){4-5}    
     & Binary     & Multi     & Binary  & Multi \\

    \midrule
    Train & 1.0 & 0.995  & 1.0  &  0.997 \\
	Val   & 1.0 & 0.996 & 1.0   &  0.998 \\
    
    \bottomrule
  \end{tabular}
  
}{%
  \caption{Accuracy of the Discriminator Network}%
}
\end{floatrow}
\end{figure}

\section{Results}

\begin{figure}[!htb]
  \centering
  \includegraphics[scale=.4]{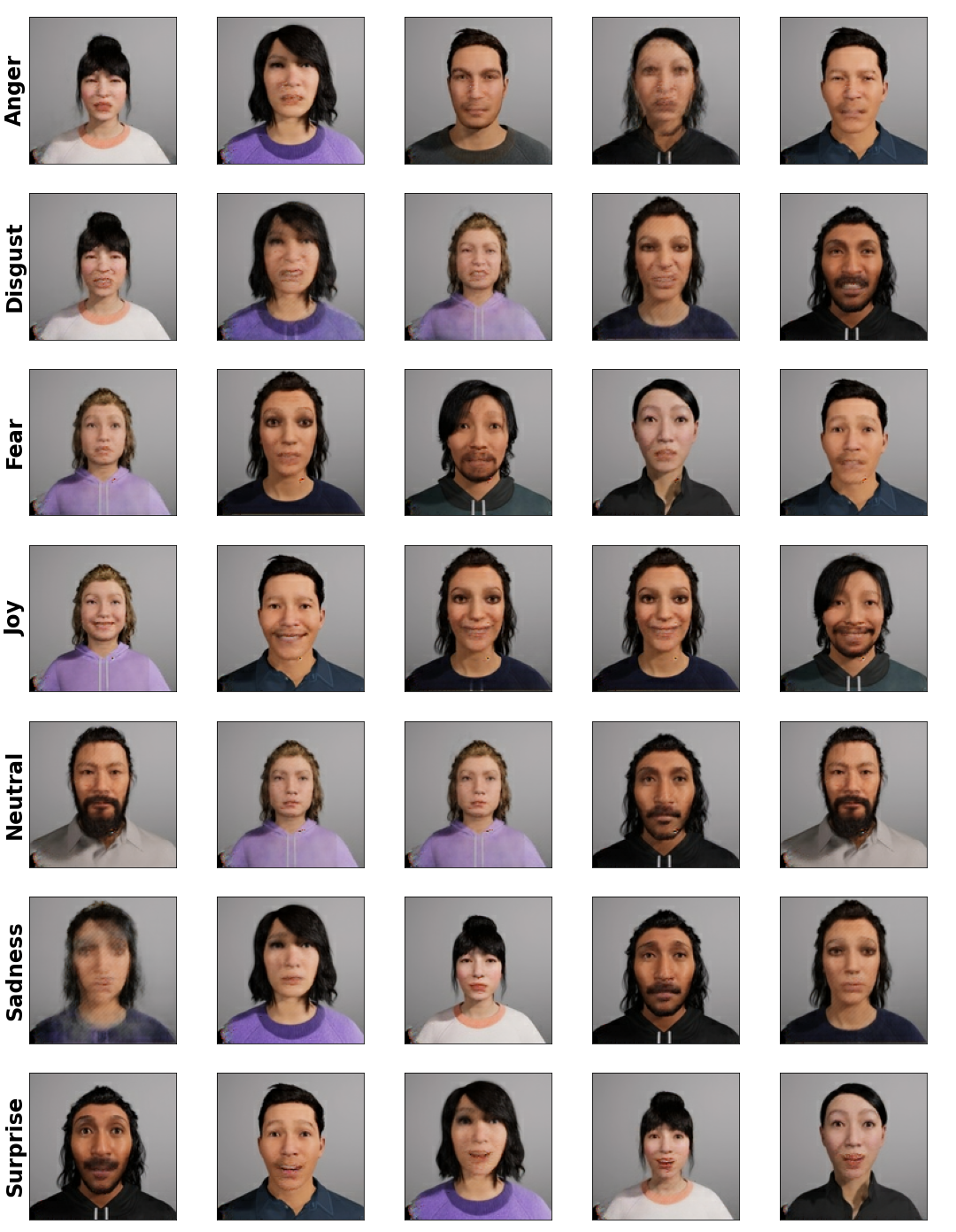}
  \caption{Images generated using Random Sampling in the Latent Space}
  \label{fig:results-random}
\end{figure}

The modular construction of GAN network makes it feasible to use each component (i.e. encoder, decoder and discriminator) separately once it has finished the training phase. The decoder when given a random vector sampled from the normal distribution along with a desired affect vector, generates an image of a Meta-Human exhibiting the desired expression. The results of the decoder network when used with random input can be seen in the figure \ref{fig:results-random}. The randomly generated images are distinct and have high variance in terms of both the character identity and expression within the character identity. This means that the network doesn't suffer from mode collapse or over-fitting problems. \\

\begin{figure}[!htb]
  \centering
  \includegraphics[scale=.25]{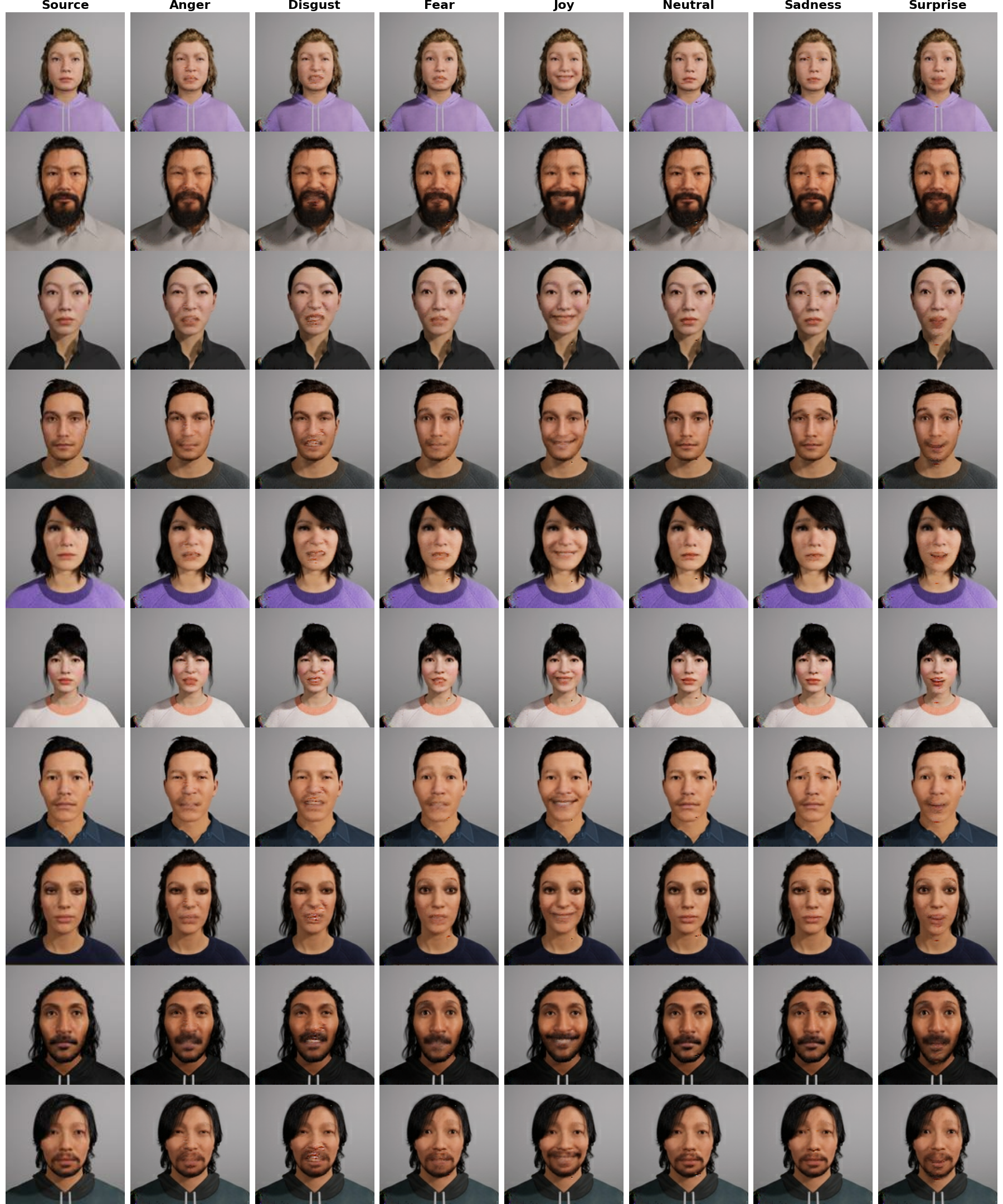}
  \caption{Results of various affect transformations for each character identity}
  \label{fig:results-reference}
\end{figure}

We analyze the results further and take a set of neutral images of each Meta-Human as a source image and then project them into latent space by encoder. The latent vectors obtained as a result, along with a set of desired facial expressions, is input into decoder network. The output of such affect transformations for the identity in the source image can be seen in the figure \ref{fig:results-reference}. It can be seen that each source image with neutral expression has been accurately transformed into the desired expression. \\

\begin{figure}[!htb]
  \centering
  \includegraphics[scale=.3]{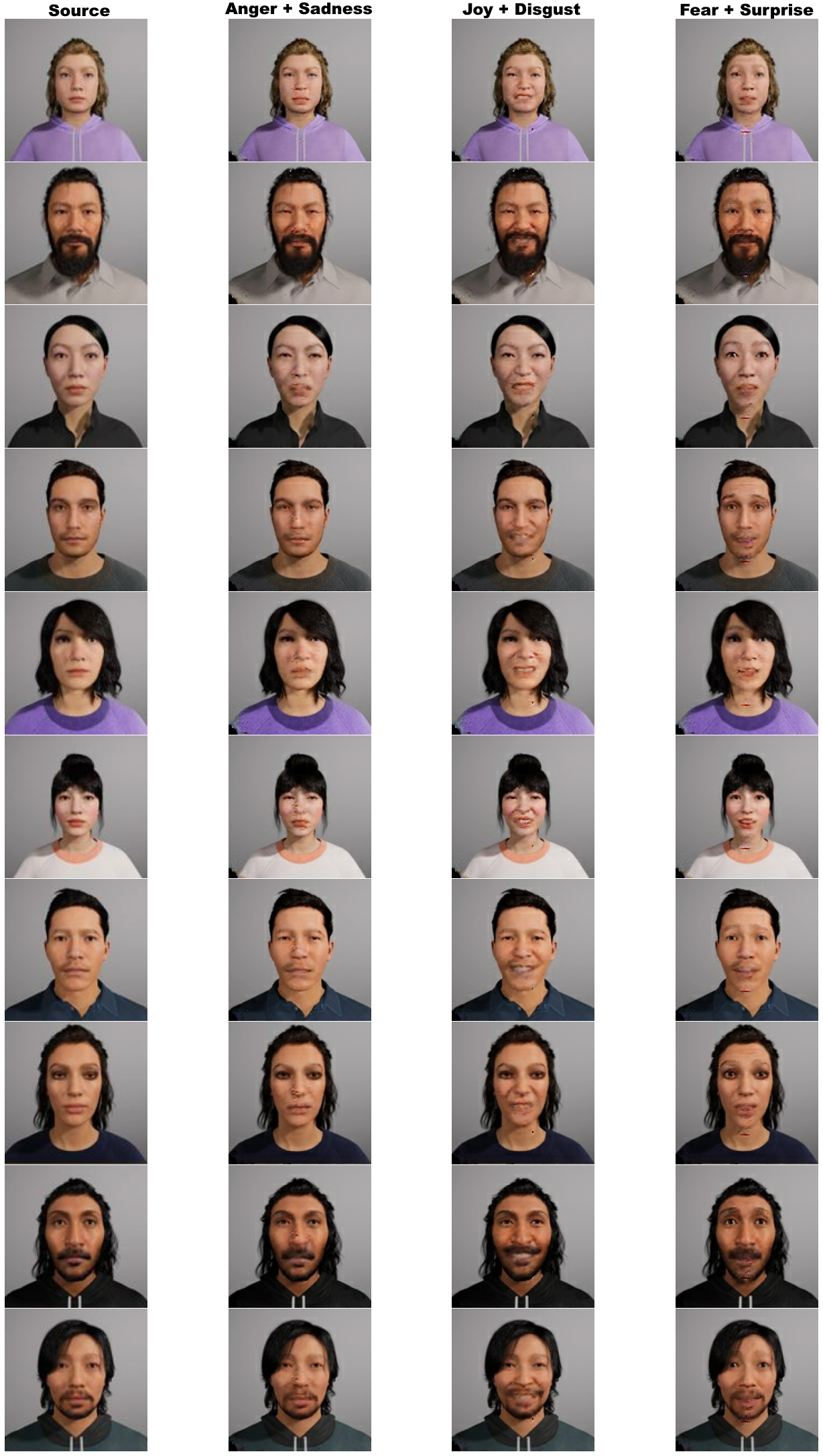}
  \caption{Image Generation for complex Expressions}
  \label{fig:results-multi}
\end{figure}

The model is robust at generating facial expressions of a particular Meta-Human identity with control for both the variations in character identity and the expression that an identity exhibits. However, We also try to generate complex expressions by combining the simple expressions in the latent space. We project a given character with a neutral facial expression into the latent space and then give a hybrid affect condition as a target expression. There are three complex expressions which we try to generate: anger+sadness, joy+disgust, and fear+surprise. The results of complex expression transformation experiment can be seen in the figure \ref{fig:results-multi}. The subtleties of facial expressions make it hard to generate complex expressions, however, the network generates a set of complex expressions effectively. Considering the fact that network has not been trained on the complex expressions, the results are promising. The results can further be improved by optimizing on  complex expressions with a few shots learning. It might also be beneficial to optimize for $\lambda$ for the complex image generation which has been chosen manually in order to generate complex expressions. The quantitative results of the discriminator network on training as well as test set are also recorded and can be seen in table 1. 

\label{sec:results}

\section{Conclusions}
\label{sec:conclusions}
In this article, we have proposed the Facial Expression Generation Network for Meta-Humans (FExGAN-Meta). A large collection of Meta-Humans image dataset is prepared by systematically placing 10 3D Meta-Humans in a studio setup and a series of facial expressions are recorded. We evaluated and verified that the FExGAN-Meta works robustly and generates the Meta-Human images with desired facial expressions. The transition from images of Meta-Human to real human images should be quite straight forward and may or may not require additional training depending on the context of the real world images in which they have been shot. 

\clearpage
\bibliographystyle{unsrt}  
\bibliography{references}  

\end{document}